\documentclass[letterpaper, 10pt]{article}
\usepackage[utf8]{inputenc}

\title{From Language Games to Drawing Games}
\author{Chrisantha Fernando \and Daria Zenkova \and Stanislav Nikolov \and Simon Osindero}
\date{DeepMind London, August 2020}

\usepackage{graphicx}

\begin{document}

\maketitle

\begin{abstract}

We attempt to automate an artistic process by inventing a set of drawing games, analogous to the approach taken by emergent language research in inventing communication games. A critical difference is that drawing games demand much less effort from the receiver than do language games. Artists must work with pre-trained viewers who spend little time learning artist specific representational conventions, but who instead have a pre-trained visual system optimized for behaviour in the world by understanding the environment's visual affordances. After considering a range of drawing games we present preliminary experiments which have generated images by closing the generative-critical loop.

\end{abstract}

\section{Introduction}

What is a drawing? Sadly, no other animal represents the world with language or drawing. Early examples of drawing date back to 60,000 years ago \cite{hoffmann2018u}, though red pigments for mark making are already found 200,000 years ago in the middle stone age \cite{ochre}. ``The first man to make a mammoth appear on the wall of a cave was, I am confident, amazed by what he had done'' writes Gibson, because they had discovered that by means of lines they could delineate something \cite{gibson2014ecological} p263. What allowed humans to learn to create (visual) abstractions, e.g., the Western child's human stick figure, the Australian aboriginal top-down projections of people seated around a fireplace, the Egyptian formalism for representing things in orthographic projection with multiple station-points, and with social dominance relations transformed into size differences, or the 16th Century Japanese affine projections that have a birds-eye viewpoint? Making our own abstraction creating (drawing) machines is one way to find out the answer. A wonderful start was made by Harold Cohen's abstract drawing programs ``Aaron'' \cite{cohen}. Aaron and Harold produced beautiful and interesting abstracted drawings that looked as if they had been made by a human alone. But it had no learning, was not conditioned on looking at the world, and was an entirely hand designed production system (a complex set of hierarchical rules for drawing).   \\

Artists have invented visual abstractions over historical timescales, for example, various perspective projections for transforming 3D things into 2D pictures. A picture is an interpretation of what it portrays. Some object property or relation, of the visual scene (e.g. the object identity or relative location) is first apprehended or isolated in the mind, then represented in the drawing, leaving out a whole lot of other unnecessary features which are of no pragmatic interest at the time \cite{arnheim1997visual}. A picture preserves a subset of the critical features of the image necessary to represent the desired property, e.g. its occluding edges, but ideally no confusing other. A lawful transformation links the image to its drawing. Consider the following examples: a child's drawing of a human figure or tree as a collection of circles, ovals and straight lines, chronology shown in a comic book, shading shown by cross-hatching. \\

The process of visual abstraction is analogous to the process of linguistic abstraction by which words are invented, but with one crucial difference. Whilst the word cat does not itself resemble a cat, and neither does the word cat sound like a cat (i.e. it is not onomatopoeic), the picture of the cat will resemble the real cat visually if it is to be understood universally. Resemblance here means that a pre-trained classifier for cats will give high probability to a cat drawing without needing further training on cat drawings. We call a representation that requires no further special training from the viewer to understand its referent, a veridical representation. On the other hand, some representations of cats are so abstract that they cannot be understood unless this pictorial `language' has been previously learned. In general, what makes visual depiction different from language is that in a lot of (but not all) cases of visual depiction it is possible to understand what the picture means without having to have any special training for the viewer other than the universal visual experience we all have. This is almost never the case in language where it is necessary to have heard that particular language very often before one can understand it, although even in this case there is deep multi-model abstraction as shown in the Kiki effect \cite{peiffer2019cerebral}, and language similarities which allow some transfer between languages. \\

\begin{figure}[h!]
\centering
\includegraphics[scale=0.3]{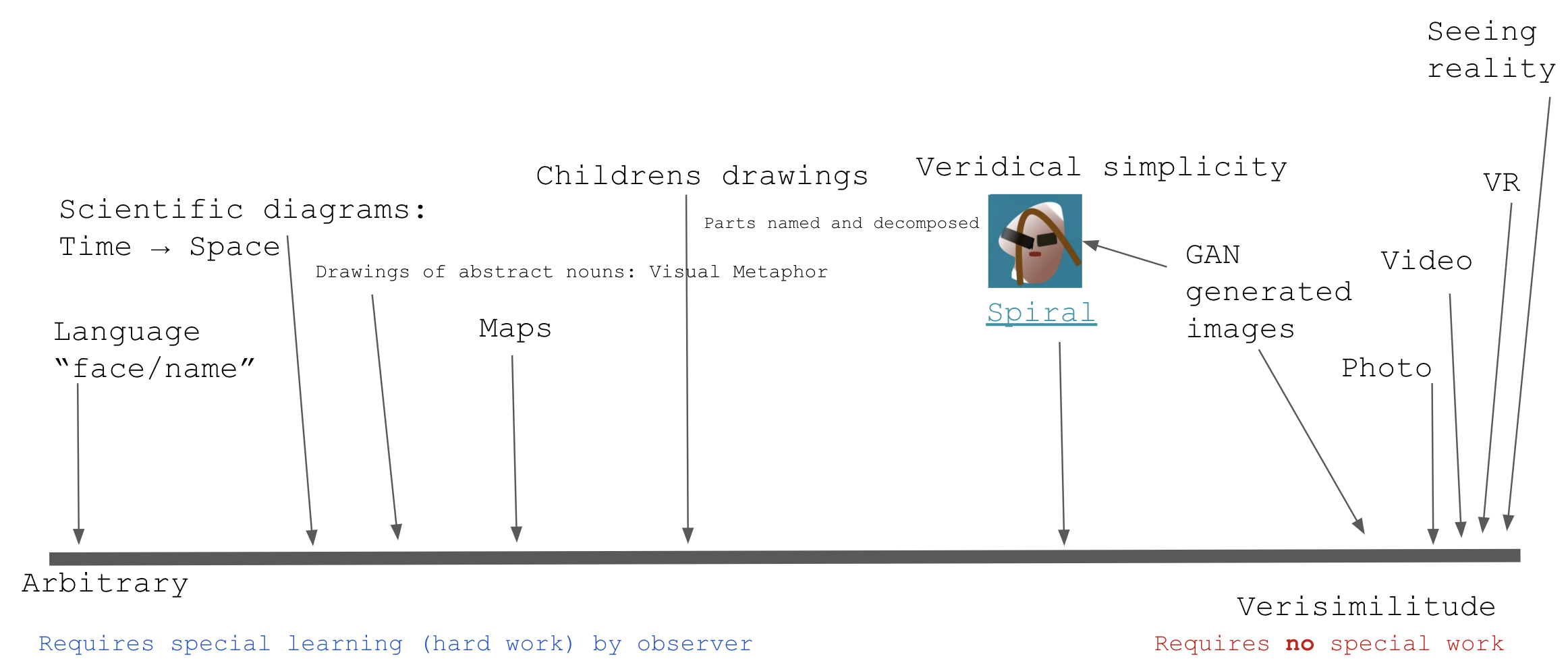}
\caption{A linear abstraction about abstraction.}
\label{Figure1}
\end{figure}

Pictures we draw can have various extents of veridicality, some verge on perfect photorealism but others are entirely arbitrary (linguistic) symbols. In the middle we have various kinds of visual abstraction, see Figure \ref{Figure1}. At one extreme, we can attempt to represent abstract nouns using drawings, e.g. we can draw democracy (cohesion with individuality), youth (upward growth), marriage (fitting together), or past, present and future, if we are asked to \cite{arnheim1997visual}. In these cases, the drawing representation is very metaphorical and the relations between the abstract noun and the drawing are more distant, but a viewer may be able to at least guess what is represented. Some scientific diagrams, e.g. Cartesian graphs, are an example of abstracted representations where considerable extra learning is needed to understand their meaning. A graph of the stock market going up does not resemble a photograph of a stock market in which stocks are going up. How does this abstraction ability arise? And what are the underlying algorithmic mechanisms? How hard is it to discover a new kind of lawful transformation between things and drawings, i.e. a new law of visual representation? What are the constraints on this lawful transformation that allow them to make sense to us? What kinds of transformation wouldn't make sense to us and why? \\

\begin{figure}[h!]
\centering
\includegraphics[scale=0.45]{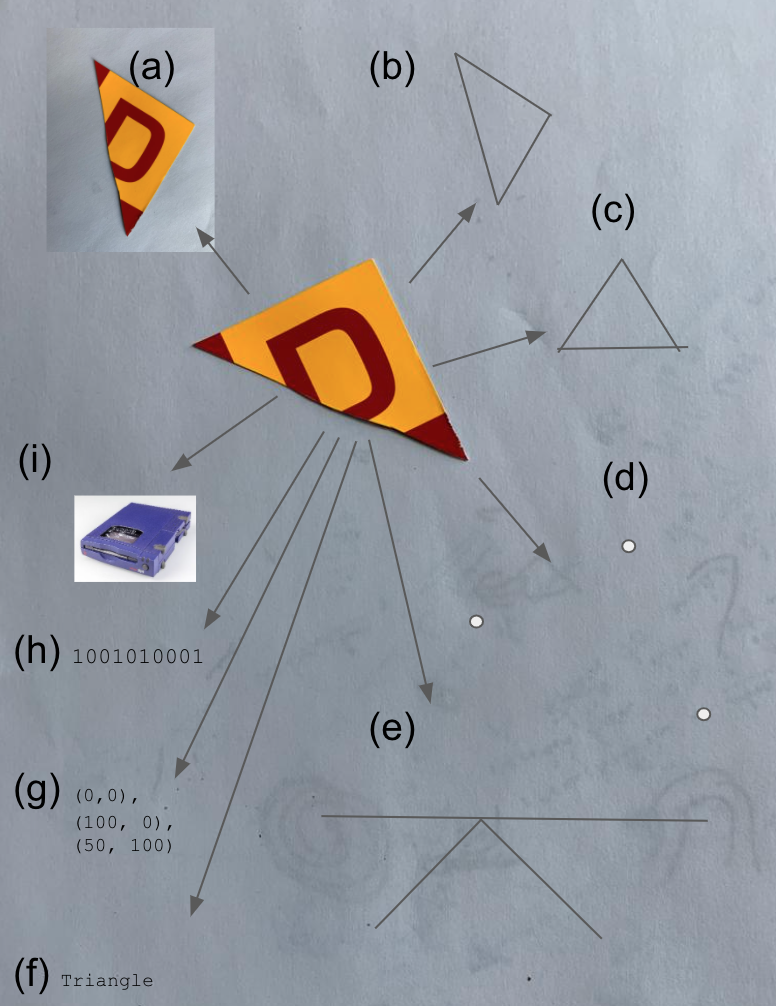}
\caption{Some ways of representing a triangle.}
\label{triangle}
\end{figure}

As a concrete example let's consider the ways in which we could represent a triangular shaped object in the world, see Figure \ref{triangle}. The first extreme is photography, you simply take a photo (a), and so each point in the scene A is copied to a pixel in picture A. Another approach is to draw the outline of the triangle as three straight lines (b), which superimpose and are at `exactly' the same relative positions as the edges of the original triangle. While this convention of using straight lines is being used, the colour, internal pattern, position, orientation, or size, or any combination of these properties of the triangle may be irrelevant for the drawing (c). In more extreme cases, the shape, i.e. the 3 lengths and angles of the triangle may even be irrelevant, in which case any three straight lines making an enclosed shape could be drawn. More minimally one might just draw 1 circle at the position of each of the vertices of the triangle (d). Alternatively one might draw the lengths of the sides as straight lines emanating from a central point, with the angles of the triangle preserved as angles between the lines emanating from the central point (e). This representation may be opaque and require some learning by the observer, but once the mapping is understood, a human would be able to generalize to other examples of triangles represented in this way quite easily.\\

At the other extreme, one may write the word triangle (f), or the x, y coordinates of the vertices of the triangle on a Cartesian axis (g), or the equations of the three line segments of the triangle, or the binary string that represents the image in memory (h). One might compress the image using some compression algorithm and represent the image as the file so produced (i). What distinguishes these representations from the previous visual abstractions? One critical way they differ is that some can immediately be understood by most humans and some cannot, without further knowledge. Understanding the gzipped file will be impossible by eye or hand. However, in all the examples both visual and non-visual above, once the lawful representation method is understood, the viewer can generalize to new triangles. Even in the case of the gzip, if a gzipping and gunzipping machine is available, then the viewer could understand any triangle represented in this way. \\

There is a kind of communication, which allows no generalization at all, and that is arbitrary non-systematic communication where for any instance of image which we wish to represent, we produce a random symbol or mark, which has no lawful relation to the image at all. It is completely arbitrary. A triangle at one position may be represented by a turtle, and another triangle by the letter 'w' and so forth. This silly method of representation would result in a complete inability to generalize beyond the training set of images shown, and as such would not be a very useful way of representing triangles going forwards. There is no smoothness between the representation and the thing which is being represented. Small changes in the thing to be represented do not manifest as small changes in the drawing. Hence the lawful transformation we are looking for should be topographic, an ordered projection \cite{harvey2015topographic}. \\

Of course "to represent a triangle" is an ill-formed task. To code this up, we need to specify a minimal game which might be something like this. Agent A is shown image A and agent B is shown image B. Agent A must draw a drawing A such that agent B by looking at drawing A can tell if their own image B belongs to the same class as image A. Both agents get rewarded if the communication is a success. The property of the triangle which must be effectively captured by the drawing A is entirely dependent on how the classes are defined in the data set consisting of 'same' pairs  and 'different' pairs. For example, 'same' pairs of image may consist of the identical triangle, and 'different' pairs may consist of triangles in which one vertex position is uniformly randomized, or 'different' pairs may have the triangle randomly rotated. The exact nature of the training set will determine what properties of the triangle must be communicated in the drawing. If the viewer is not trained for each specific drawer, then a good drawing of a triangle A is one which can allow this pre-trained viewer classifier to tell in one shot if the triangle A is the same as triangle B by looking at the drawing A and triangle B. So a drawing game is a subset of a language game \cite{lazaridou2018emergence}, but we add to the language game the constraint that there does not need to be any (or much) learning by the viewer, other than their prior visual experience with the world. Take the example of a map. A map  exploits our pre-existing experience of looking at things from the top, or our existing allocentric neural representations \cite{banino2018vector}. If we didn't have these priors, it wouldn't be a useful visual abstraction, learning to read maps would be much harder for us, maybe as hard as learning language. An analogy of drawing exists in the auditory domain then, which is onomatopoeia and sequencing of sounds in time, i.e. auditory mime. \\

An interesting experiment is to take two images from two different ImageNet classes \cite{deng2009imagenet}, show this to the drawing agent and reward the drawing agent for making a drawing which maximizes the probability of class A and minimizes the probability of class B in pre-trained ImageNet classifier. This is closely related to the cycle consistency work of Chen et al's in `Constrastive Learning of Visual Representations' except instead of learning a latent vector we wish to learn a drawing \cite{chen2020simple}. A greater simplification is to show one image from a class and reward the drawing agent for maximizing that class probability for a viewer who observes their drawing. This would lack the contrastive element, but if the drawing agent were trained on a set of training classes then it would be fascinating to see if the agent could generalize to novel classes by the general drawing procedure learned, e.g. would it learn orthographic perspective, or the drawing of occluding edges, or shading? Would it be able to draw classes it had not drawn before, and yet be able to make them recognizable? If so it would be a general drawing agent, having had to discover methods of representation that were general to many drawable things, at least the very limited set of things the pretrained ImageNet classifier was able to recognize. Even imaginary scenes may be possible, e.g. to achieve half-fish-half-cat, one might wish that both ImageNet classes of fish and cat had high probabilities. To achieve robustness one might wish to make the drawing resemblance be high even if the drawing was viewed from many different angles. Would this result in a cubist type representational drawing with multiple perspectives being shown? Why should the drawing agent be conditioned on an instance of the class? Because it needs to generalize to new classes to be effective. The above experiment has already been done with humans. Hawkins et al observed pairs of humans who represented objects to each other using drawings. As expected, they found that over repeated interactions the drawings became more abstract and less veridical \cite{hawkinsgraphical}. \\ 

A closely related piece of work is the Perception Engine by White \cite{white2019shared}, where a stochastic search algorithm was used to optimize (unconditional on any viewed scene) a drawing to maximize a certain ImageNet class probability. White avoided the pathological false positive tendency of a single neural network trained on ImageNet \cite{nguyen2015deep} by requiring that multiple independently trained networks give a high belief in the class, and also adjusting lighting and viewpoint and hue in the drawing. He produced images that were in a sense `prototypical' \cite{wittgenstein2009philosophical} representations of the object for those neural networks that had been trained on ImageNet. The exact nature of this prototypy is a fascinating subject for investigation if we are to understand how imageNet classifiers work. One consequence of White choosing to require the classifier to view the drawing from multiple viewpoints is that it has encouraged the drawing to have it's own special kind of Gibsonian invariance \cite{gibson2014ecological}, whereby the drawing must look like the object from multiple angles simultaneously, one of the ecological motivations of cubism \cite{glassner2004digital}. In fact, Gibson proposes that such an abstractive tendency exists in childrens' drawings of objects, which highlight invariant structure and discount perspective structure, \cite{gibson2014ecological} pp78, saying ``while drawing, he may be looking at something real, or thinking about something real, or thinking about something wholly imaginary; in any case, the invariants of his visual system are resonating. The same is true of the artist as of the child. The invariant are not abstractions or concepts.'' \cite{gibson2014ecological} p266. The reason that children may more often choose to draw orthogonal projections is that they have the greatest number of invariants, e.g. angle, parallelism, shape and relative size are conserved and independent of the viewer's fixed angle in space \cite{freeman1985visual} p74. He goes as far as to say ``What modern painters are trying to do, if they only knew it, is paint invariants. What should interest them is not abstractions, not concepts, not space, not motion, but invariants.'' \cite{gibson2014ecological} p271. Can one say that such invariants are captured in drawings of objects from memory, as in the Sketch-RNN dataset \cite{ha2017neural}? Certainly this dataset does not result in high ImageNet classifier class probabilities. What exactly is the difference between invariants, abstractions, and concepts? In any case, perhaps if Imagenet classifier networks could buy art and exhibit them in museums there would be several rooms filled with these masterpieces. This would be further confirmed if our own human paintings of objects tended to give higher class probabilities than photos  of those objects shown to human observers. \\

We have already made headway in understanding how veridical representations of natural scenes can be generated. For example MONET \cite{DBLP:journals/corr/abs-1901-11390} shows that by learning visual attention masks at the same time as learning to autoencode the masked units, it is possible to discover 'objects' from a visual scene along with soft saccade like attentional processes moving over the scene to identify such objects.  With some extra work these can be clustered into object classes. Once an object class has been defined by some yet unknown abstraction process, humans can draw archetypal instances of that class, e.g. as shown in the QuickDraw dataset \cite{ha2017neural}. Using supervised learning it is possible to learn an LSTM-GMM model that can reproduce these human sketches (SketchRNN). Given a particular class of images such as faces, it is possible to train a GAN to produce a veridical drawing (Spiral++) of that class \cite{mellor2019unsupervised}, sometimes with quite abstract looking depictions of the type used by children, e.g. a mark for each of the two eyes and a mark of a nose and a mouth, which can fool a discriminator effectively. So we know that an adversarial GAN algorithm can produce stylized drawings of objects recognizable as that class. So spiral++ (GAN) is an instance of an abstraction making machine capable of producing veridical visual abstractions. \\

But no GAN can produce non-veridical abstractions. For example, if a GAN were shown visual scenes of Mount Fuji taken from different terrestrial locations, a generator would never come up with a contour map of Mt Fuji seen from above, because a discriminator could easily discriminate the tourist photos from such a map. There is an intermediate zone of visual representations which is neither photorealistic, nor entirely arbitrary like language. To learn conventions in this zone requires some extra information in addition to that which can be obtained from photographic visual images themselves. For example, to understand a map one has to imagine the world from the top, and to do so one probably has to move through the world, i.e. the representation is action dependent. A transformation has been made in the imagination of the position and distance of the observer, floating high up in space. This gives us a clue to the rules of a game which would result in drawings being made to represent things in a more abstracted way than GANs can achieve, and leads us to the seminal work of James Gibson \cite{gibson2014ecological}. \\

To address the question we have begun to train drawing agents by reinforcement learning to play a variety of `drawing games'. The hope being that playing these games will result in the origin of interesting visual representations of things in the world. All the games take an adaptationist view similar to that taken by art historian Earnst Gombrich in `The Story of Art', where new art arises as a solution to a new representational problem. Non-representational drawings can also be analyzed within an adaptationist framework, where each drawing is trying to solve a more abstract problem. This adaptationist view helps makes sense of abstract and conceptual art as well. Within our framework a particular art form is formally defined in terms of computational aesthetic reward functions (what the art is trying to achieve), and productive constraints on actions (i.e. what limits are placed on mark-making methods). We envisage the artist's task as being to invent new and interesting games to play, and to devise new and interesting constraints for playing this new game. However, in reality, the reward functions of these games may be so complex, requiring a full model of the human brain, emotion, and society, that it may be impossible to write them down concisely, or at all. This perspective allows us to ask questions such as ``How many marks did Picasso look ahead?'' i.e. to think of artistic mark making as involving constructional planning, and also ask e.g ``what was Rothko trying to maximize?''.\\

\section{Methods and Results}

\begin{figure}[h!]
\centering
\includegraphics[scale=0.25]{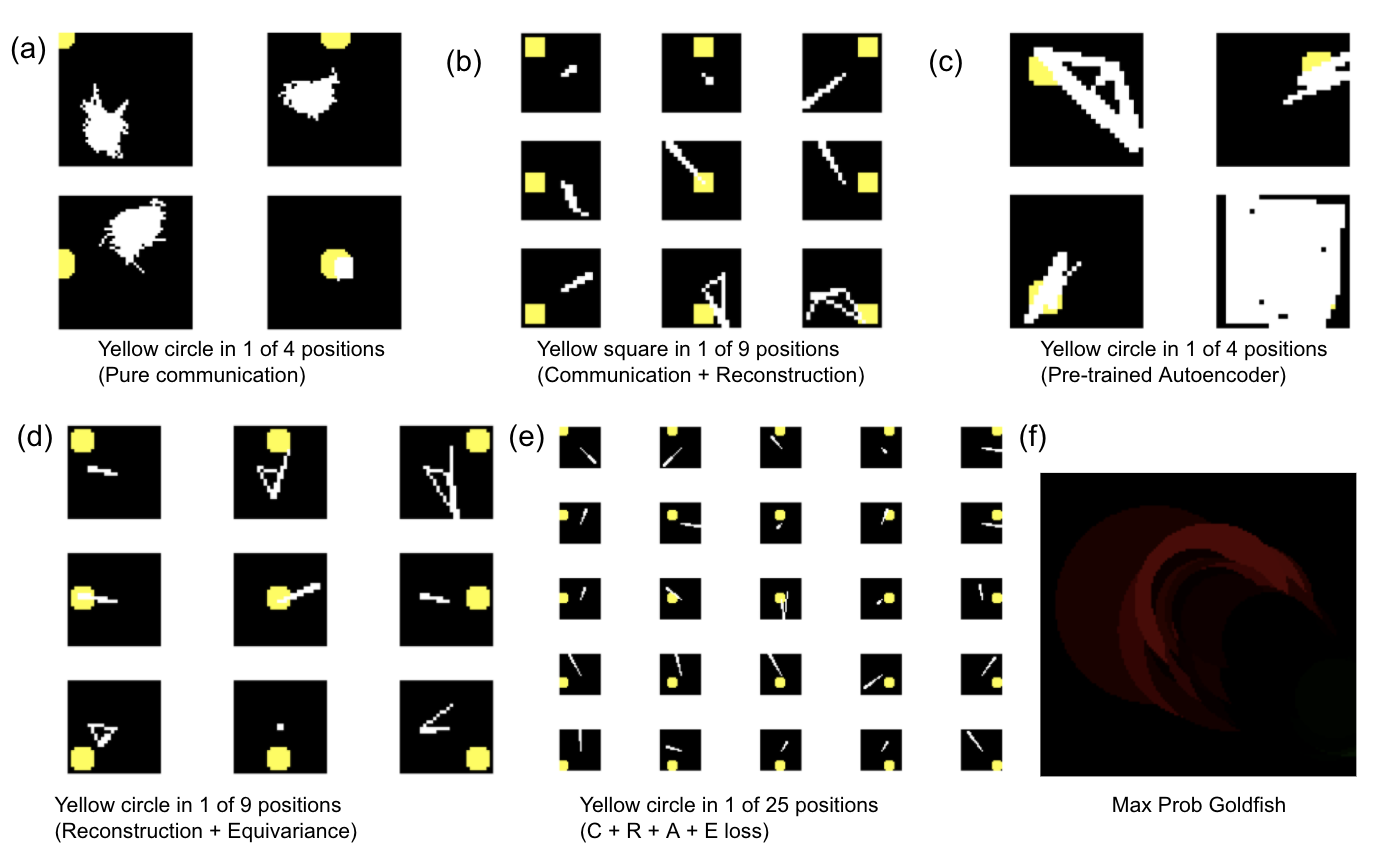}
\caption{Drawings made in 6 kinds of drawing game, see text for details}
\label{Figure2}
\end{figure}

\subsection{Pure communication game}
Our first game treats drawing as the solution to a pure communication problem, see Figure \ref{Figure2}a. Agent A observes a scene (a yellow circle at a particular position), and must make a drawing such that when agent B observes their own scene and the drawing, agent B is able to guess correctly whether the drawing is of their scene or not. A dataset consists of pairs of scenes and balanced `same` or `different` labels. Agent B is trained by a cross entropy loss for classification. Agent A is trained by reinforcement learning (A2C) \cite{mnih2016asynchronous} with the reward being the negative log cross entropy loss, which increases as agent B improves its classification accuracy. The drawing is a physical bottleneck in communication. \\

\subsection{Pure reconstruction game}

The second game, Figure \ref{Figure2}b, requires that agent A observes a scene, and makes a drawing such that a reconstructor R is able to observe just the drawing and reproduce the scene that agent A saw. R is trained using a mean squared error loss and Agent A is trained by A2C with the reward being the negative log mean squared error, such that perfect reconstruction produces higher rewards. \\

\subsection{Pre-trained autoencoder based reconstruction game}
The third game, Figure \ref{Figure2}c, requires that agent A observes a scene, and makes a drawing such that a convolutional autoencoder pre-trained on scenes, is able to observe the drawing and reconstruct the scene. The autoencoder A is never trained directly from drawings to scenes, but only on scene to scene. This is intended as a pressure for the drawing to resemble the scene (from the point of view of the pre-trained autoencoder). Agent A is again trained by A2C where the reward is the negative log mean squared error of the pre-trained autoencoder. This is the most veridical example of drawings so far. \\

\subsection{Equivariance game}
The forth game, Figure \ref{Figure2}d, requires that agent A observes a scene, and makes a drawing D. But this time, agent A also observes a transformed version of the scene and must also make a drawing D2 of that transformed scene. A mapper M network is then trained using mean squared error loss to convert D into D2dash. M is trained to minimize the mean squared error between D2 and D2dash. The negative log loss of M is then used as the reward to train agent A using A2C. This is maximized when the two drawings D2 and D2dash are identical. It is hoped that to achieve this the drawings should be a systematic function of the scene, otherwise it would be harder for M to reliably produce D2dash (resembling D2) from D. At the same time a reconstructor R is trained to be able to reproduce the corresponding scene from D and D2. The reward for the agent is a linear function of the sum of the two negative log reconstruction losses and the negative log loss of M (which is the equivariance loss E) \cite{lenc2015understanding}. Without the reconstruction loss a trivial solution would be to produce always the same drawing. Some systematicity can be seen in the top row, however, it is not clearly shared over all locations of the yellow circle. \\

\subsection{Mixed loss drawing games}
A fifth game combines all of the above losses, Figure \ref{Figure2}e to solve a slightly harder representational problem where the circle can be in one of 25 positions.\\

\begin{figure}[h!]
\centering
\includegraphics[scale=0.3]{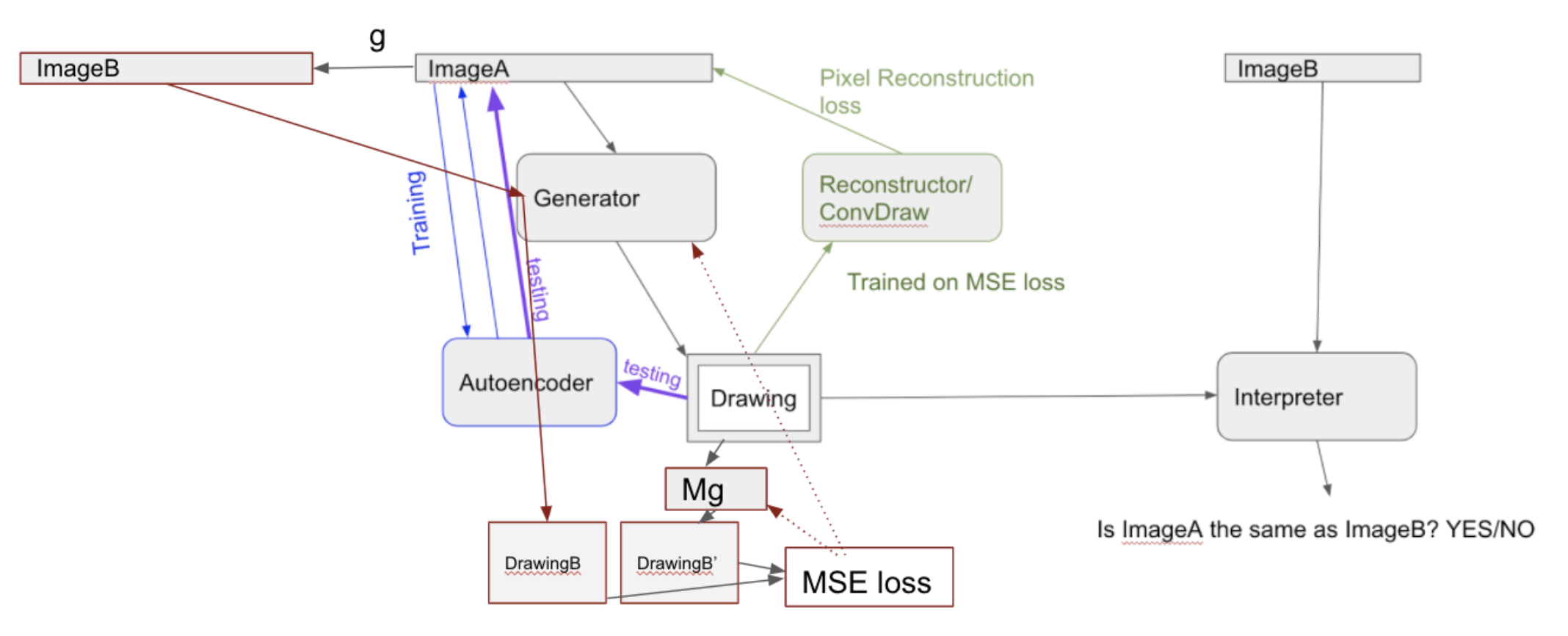}
\caption{Architecture of networks used in drawing games \ref{Figure2}a-e.}
\label{Figure7}
\end{figure}

In all the games we see that only by using the pre-trained autoencoder loss was there any tendency to verisimilitude. Figure \ref{Figure7} shows the architectural outlines used in the first five drawing games shown in Figure \ref{Figure2}a-e.\\

\subsection{Conditional ImageNet classification game}

A sixth very different game, Figure \ref{Figure2}f, extends game \ref{Figure2}c and requires an agent to observe a random instance of an ImageNet class and draw that class such that a pre-trained ResNet50 \cite{he2016deep} gives high probability to that class. In this case unfortunately we have not yet been able to train a general image conditional drawing agent, because the drawing agent always learned to draw an unconditional image of the class `from the imagination', rather than drawing what it sees. We intend to extend the DRAW framework for image conditional drawing \cite{gregor2015draw}, training it not to autoencode but to produce the outlines of objects from photos, and then subsequently optimize its parameters for maximizing an ImageNet class probability. Because our drawing agent did not learn to condition itself on the class image, it was not able to generalize to drawing other classes. Our drawing agent has not had any prior visual experience independently of suddenly being asked to draw a convincing image for a ResNet. This is not the case with a visual human artist. The action space of the drawing agent is to produce a circle of a given color at a given position, a richer action space may be helpful. Also, we note that the process of hand-eye coordination in depicting by drawing is not yet understood. Artists look back and forth between the scene and the depiction, \cite{freeman1985visual} p42 with brief glances with one or two fixations, they may half close their eyes, make measurements, learn forward, stand back, choose materials, set perspective, use measurement techniques, squint, look for shadows, drawing without looking directly at the paper, explore the lines and focus on specific aspects of the drawing or scene, etc.. What is exactly happening in this process? Is the artist making same/difference judgments between the scene and the depiction, and how can this be modelled in a conditional drawing agent? \\

\begin{figure}[h!]
\centering
\includegraphics[scale=0.35]{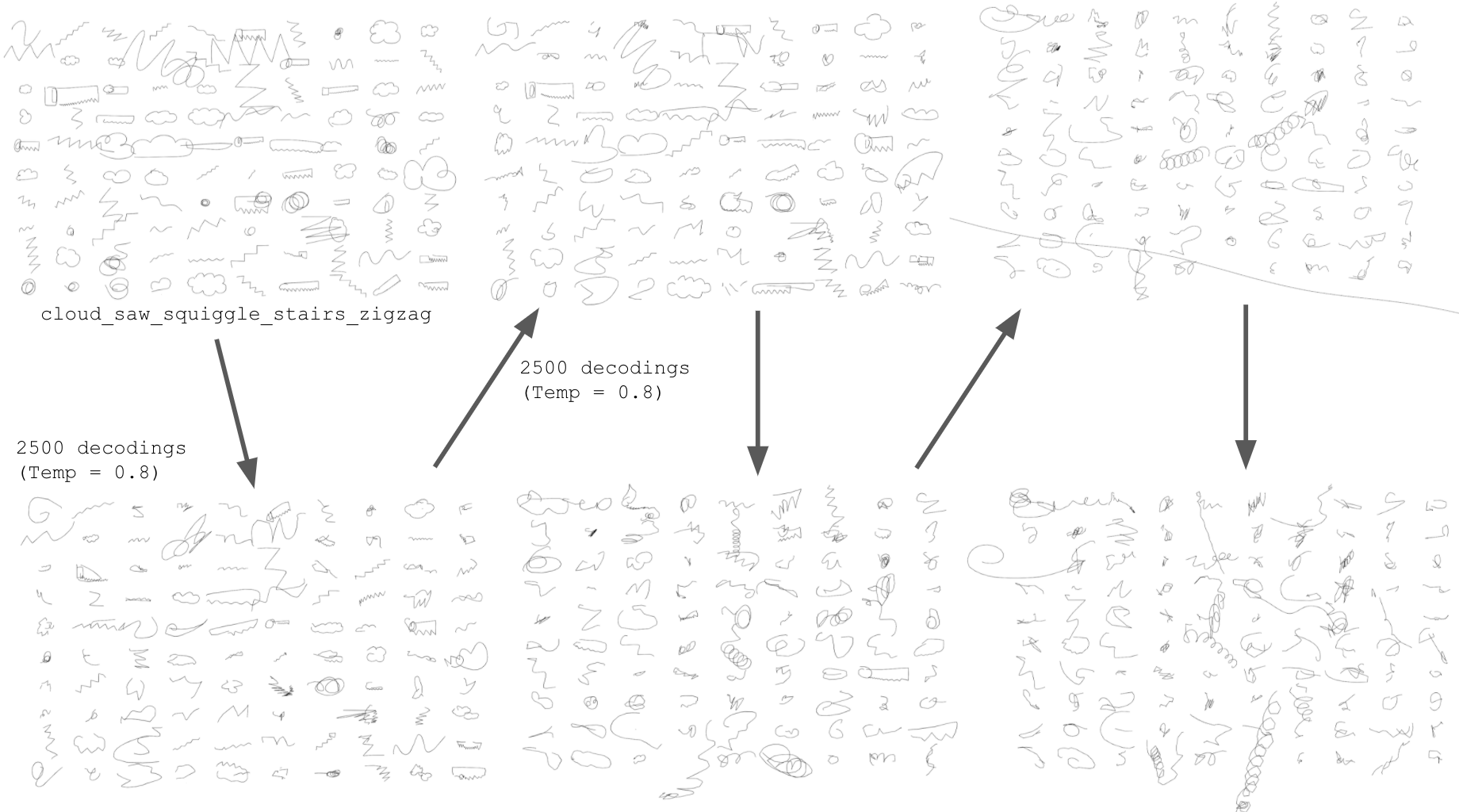}
\caption{Sketch RNN trained on 5 classes, auto-training itself based on its own productions. Over 5 iterations, very pleasing human doodly squiggles arise. The resulting network with its squiggly prior can then be trained in an RL setting.}
\label{Figure3}
\end{figure}

One solution to the problem of how to provide more general visual experience for the drawing agent is to use a pre-trained network with some visuomotor abilities, and subsequently train it to optimize certain representational goals through drawing. As a start one might take sketch-RNN (which is an LSTM-GMM) pre-trained on a subset of similar object classes and re-train it on its own drawings, which will result in an interesting degeneration (and exaggeration) of drawings into abstract forms, see Figure \ref{Figure3}. These human-like drawing priors can then be modified to achieve a task. \\

\begin{figure}[h!]
\centering
\includegraphics[scale=0.5]{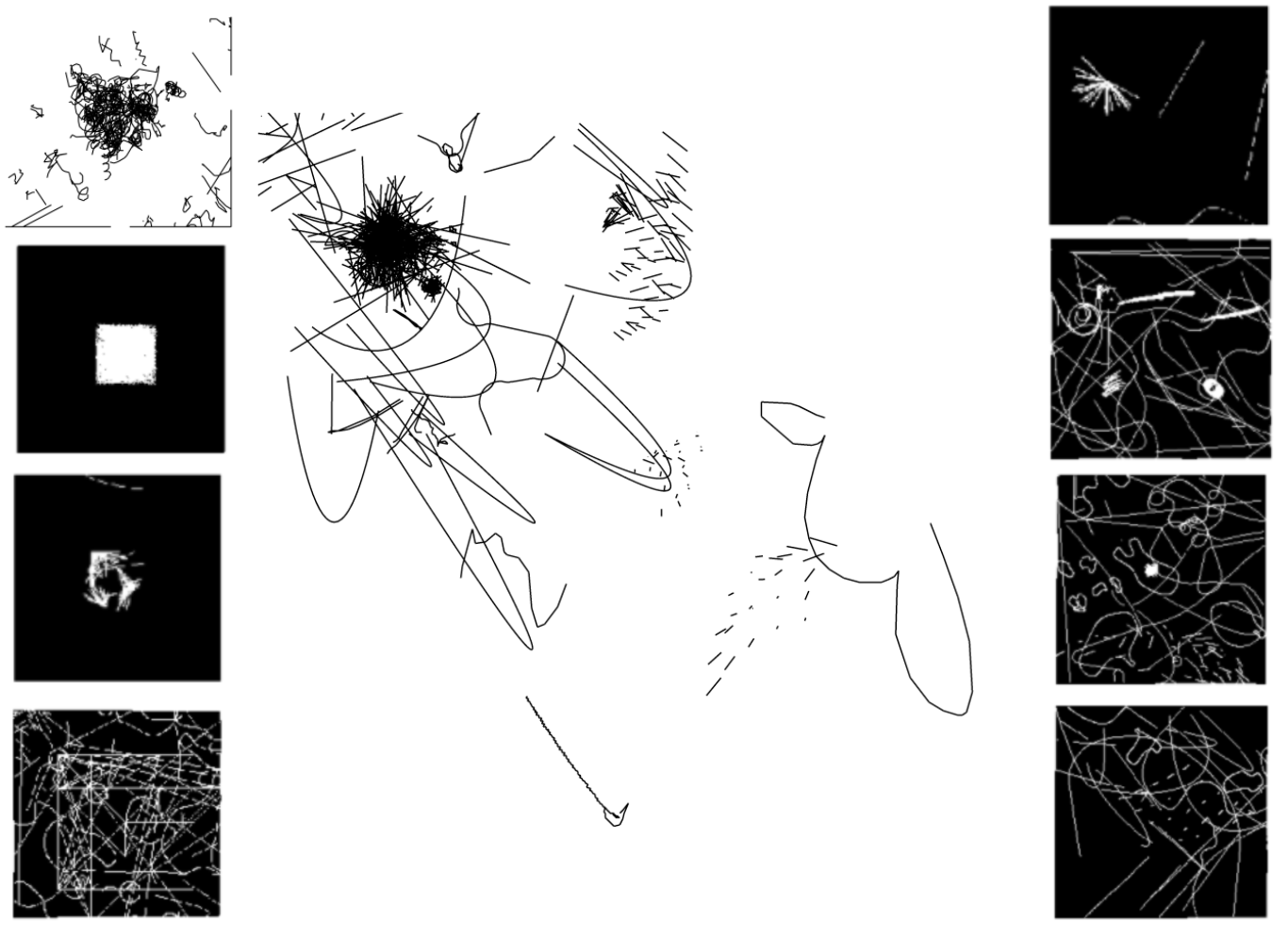}
\caption{Some drawings optimized using MCTS search over a space of high level drawing module functions, selected using MAP-Elite with a variety of computational aesthetic fitness functions, see Appendix A. }
\label{Figure4}
\end{figure}

\subsection{Abstract drawing games}

\begin{figure}[h!]
\centering
\includegraphics[scale=0.51]{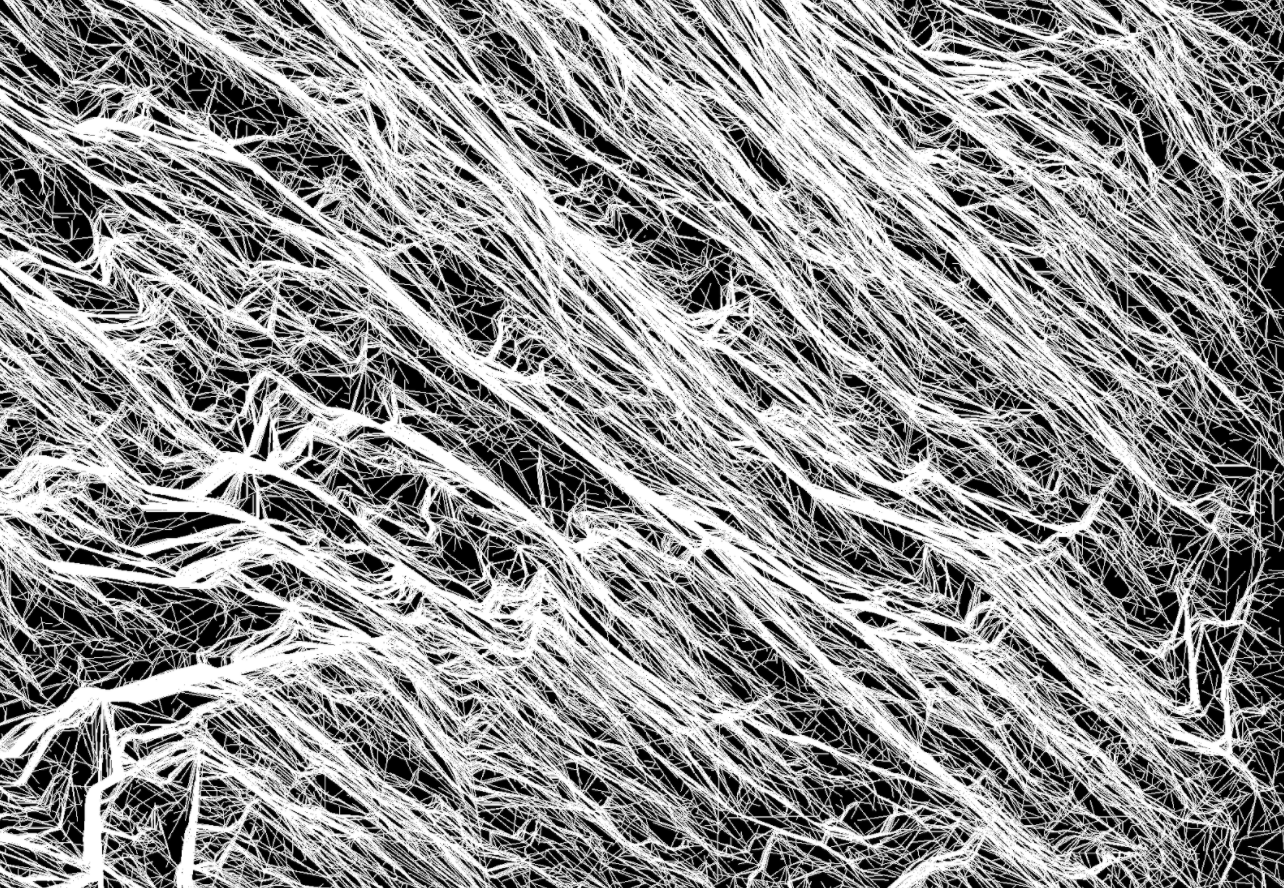}
\caption{An image produced by a simple hand designed rule based system which draws random straight lines, but turns them when they meet another line to the average its own and the other line's angle.}
\label{Figure5}
\end{figure}

\begin{figure}[h!]
\centering
\includegraphics[scale=0.6]{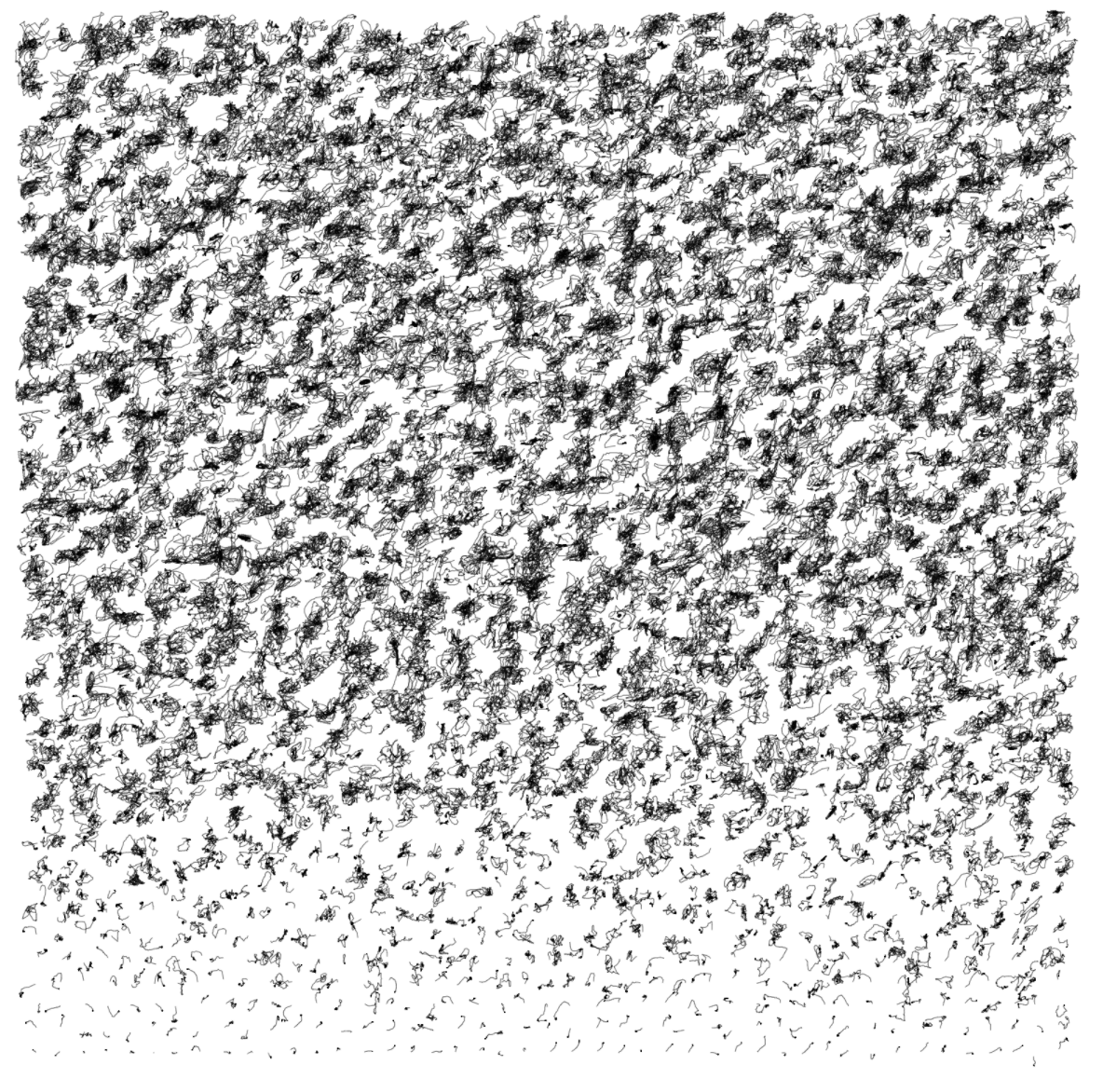}
\caption{LSTM evolved to fill in a circle. Bottom left to top right are shown sequential samples of fitness evaluations.}
\label{Figure6}
\end{figure}

A seventh even more different drawing game focuses on the production of non-representational art. Figure \ref{Figure4} shows some images produced by a very different set of non-representational drawing games in which agents are evolved to optimize computational aesthetic reward functions, using MCTS rollouts of depth 5, with selection, to choose the most rewarding sequence of marks. The actions over which MCTS acted were themselves complex pre-specified drawing modules. MAP-Elite was used to preserve a diversity of drawing solutions, with the dimensions also being other computational aesthetic fitness functions. Appendix A lists briefly the set of measures used and the generative procedures implemented. In these games there is no attempt at representational art, the functions are entirely abstract such as for example trying to achieve a scale invariant fractal dimension. Let us assume that Jackson Pollock in making his drop paintings was playing a game and succeeded in scoring highly on that game. We can work backwards and ask what game he was playing. His drip paintings exhibit a scale invariant fractal dimension. Pollock appears to have isolated the underlying fractal dimension of human movement (at the wrist and body) without any extraneous representational information, even prior to the mathematical discovery of fractals \cite{jones2009drip}. This perhaps explains why his art is interesting. One property may be maximized while simultaneously trying to maximize another, e.g. filling in a central circle, or maximizing Phog hierarchical self-similarity \cite{amirshahi2012phog} of the image whilst simultaneously reducing the entropy of the image \cite{rigau2007conceptualizing}. These juxtaposed constraints may sometimes result in interesting solutions. The artist here is modelled as a generative planning module interacting with a self-critical but diversity maintaining aesthetic module: MAP-Elite was used to maintain diverse generators along these aesthetic dimensions \cite{mouret2015illuminating}, and it may model the artistic practice of allowing serendipitous playful discoveries, rather than focusing on one objective alone. In these experiments human control was greatly relinquished and accordingly it became much harder to make interesting images. In fact, we found that some of the images produced by just a single generative module are often the most pleasing, sometimes even without the need for any explicit aesthetic evaluation, e.g. see Figure \ref{Figure5} which arises from a simple rule based system which draws random lines but which changes the angle of the current line to the average angle of itself and the line that it meets, resulting in organic thread like forms arising over time, and \ref{Figure6} which shows the result of using a genetic algorithm to evolve an LSTM module to fill in a central circle with no more complex additional constraints. Here our own implicit aesthetic functions are used in the evaluation of these images, but it is not known what yet fully what they are. Certainly such understanding would improve recommender systems. Figure \ref{Figure6} shows the evolutionary history of marks made over sequential fitness evaluations. In my opinion, the search for explicit hand designed aesthetic fitness functions described by equations may be misguided, a conclusion also later reached by Stiny \cite{stiny2006shape} despite his earlier efforts 50 years ago to come up with such functions \cite{gips1975investigation}. It is likely that a greater promise comes from a more direct use of data itself in the construction of aesthetic functions. A further argument against the search in equation space for computational aesthetic functions comes from the possibility that what is good taste in art may be learned from data biased by which works have high social status currently in the art world \cite{perry2014playing}. \\  

\section{Conclusions}

We have only scratched the surface of all the drawing games which could be played by and between agents. What if the viewer is not ImageNet but an reinforcement learning (RL) agent? This RL agent needs to solve a problem, e.g. navigate to some location. The drawer knows the location but must represent it by drawing something in the world that the RL agent can see. Will they draw a map? Under what pragmatic conditions will a map arise? If a viewer must predict when to gather reward, and the drawer knows the reward pattern, will the viewer ever learn to draw a Cartesian graph? If the viewer has emotions (drives) in an RL game that makes it relax, run away, be hungry etc., then the drawer's job would be to manipulate the viewer into one of these emotional states by drawing something. What will it draw? Perhaps if the RL agent plays Atari games it will draw something to make the viewer run away, a supranormal stimulus of a skull, making the agent jump right with very high probability? OK, I admit, `go right’ isn't quite an emotion, but the idea is the same. In one extreme case if a drawing maximizes the probability of a viewer buying the thing the drawing represents, then we have achieved optimal advertising. An artwork is fully embodied in behaviour and hence emotion, not just classification and discrimination. \\

Whilst deep learning has made great strides in object recognition, we still know very little about the neural architectures needed for relational thinking and abstraction of the type which causes constructional apraxia in lesions of the Parietal Cortex \cite{summerfield2019structure}, and which is needed for understanding of visual concepts such as proximity, similarity, alignment and closure. All these concepts are important for artistic composition. General scene understanding at the level of visual affordances will be required for the kinds of visual metaphor necessary for drawing abstract nouns, and also to make sense of scenes such as ``the last supper'' in which it is the relation of objects that matters; that Jesus is in the middle \cite{fleuret2011comparing} of a alignment of apostles, and that they have potentials for reaching bread and wine etc.. The scene is rich with possibilities for action. ImageNet classifiers know nothing of these relational concepts. Once such artificial relational viewers exist then the work of White's perception machines can be extended to more complex scenes. In fact, many of the rules of Harold Cohen's Aaron explicitly hand coded such concepts, such as detection of overlap and occlusion, balance of composition, etc.. Children also notice certain kinds of invariant in their drawings such as whether a line is straight or curved, that a line has a beginning and an end, that lines can change direction with zig-zags or continuously, that objects can be lined up, that a line can delineate a closed form, that lines can intersect, or can be parallel \cite{gibson2014ecological} p264. These are the kinds of properties that Paul Klee describes in his famous but incomprehensible Pedagogical Sketchbook \cite{klee1953pedagogical}. Also they unconsciously and spontaneously appear in the visual equivalent of mind wondering; doodling, in which without much thought we generate and explore visual patterns \cite{smallwood2015science}. But virtually all of these kinds of picture understanding are so far lacking in neural networks. \\

The next step for us is to attempt to close the creative loop again by producing an effective trainable image conditional drawing architecture and an effective parietal lobe gestalt `abstractor' which can discover novel dimensions of relational concepts for the drawing agent to enjoy generating extra-ordinary instances of. How should such an device for relational scene understanding be trained?  \\

\section{Appendix A}

Computational aesthetic reward functions used. 1. Fill central circle. 2. Mean bilateral entropy 3. Negative entropy. 4. Phog self-similarity 5. Phog complexity, 6. Region entropy, 7. Number of regions, 8. Felzenzawalb segments 9. ImageNet entropy, 10. Distance from a power spectrum target. \\

The complex drawing modules are briefly outlined as follows. 1. Random distributions of line. 2. Random lines with specific behaviours when intersecting another line. 3. Pens with local allocentric views controlled by neural networks (turtles). 4. Generators of random smooth bezier objects. 5. Random compositional pattern producing networks that output grids of lines with lengths and directions. 6. Simulated pen holding robot arms that produce random oscillatory motions. 7. Random LSTM agents. 8. Random LSTM-GMM agents using the same archiecture as SketchRNN. 9. Pre-trained and self-trained sketchRNN agents.

\end{document}